\documentclass[10pt,conference,a4paper]{IEEEtran}
\usepackage{booktabs}
\usepackage{graphicx}

\usepackage{xparse}
\usepackage{amsmath}
\usepackage{amssymb}
\usepackage{color}
\usepackage{pgfplots} 

\usepackage[font=small]{caption}

\usepackage{notation_memmesheimer}
\usepackage{hyperref}
\usepackage{lipsum}
\def\andothers{et al.\ }
\def\figname{Fig.\,}
\def\fignamelong{Figure\,}
\def\secname{Section\,}
\def\tabname{Table\,}

\ExplSyntaxOn
\NewDocumentCommand{\storereviewer}{mm}
  {
   \bcp_store_data:nn { #1 } { #2 }
  }

\DeclareExpandableDocumentCommand{\getreviewer}{O{1}m}
 {
  \bcp_get_data:nn { #1 } { #2 }
 }

\cs_new_protected:Npn \bcp_store_data:nn #1 #2
 {
  \seq_if_exist:cF { l_bcp_data_#1_seq } { \seq_new:c { l_bcp_data_#1_seq } }
  \seq_set_split:cnn { l_bcp_data_#1_seq } { } { #2 }
 }
\cs_generate_variant:Nn \seq_set_split:Nnn { c }

\cs_new:Npn \bcp_get_data:nn #1 #2
 {
  \seq_item:cn { l_bcp_data_#2_seq } { #1 }
 }
\ExplSyntaxOff

\def\ntuoneshotimpro{5.6}
\def\ntuoneshotimproreduced{3.7} % with just 60% of the data we still are better than the baseline
\def\ntuoneshotimprosecond{8}
\def\ntuoneshotresult{50.9}
\def\ntuoneshotresulttriplettripletmarginminer{50.6}  %triplet margin loss only, triplet margin miner, batch size 2000 pretrained resnet18
\def\ntuoneshotresultcrossentropytripletmarginminer{40.4}
\def\ntuoneshotresulttripletcrossentropytripletmarginminer{50.5} 
\def\ntuoneshotresulttriplet{52.2} % Triplet margin loss only 0.4241689933098035 batch size 2000 pretrained resnet18
\def\ntuoneshotresultcrossentropy{40.4} % Cross entropy 0.43080651108886897 batch size 2000 pretrained resnet18

\usepackage{pgfkeys}
\pgfkeys{
 /points array/.is family, /points array,
 .unknown/.style = {\pgfkeyscurrentname/.initial = #1},
}

\newcommand\resultsstorage[1]{\pgfkeys{/points array, #1}}
\newcommand\results[1]{\pgfkeysvalueof{/points array/#1}}

\resultsstorage{apsr20 = 29.1, apsr40 = 34.8, apsr60 = 39.2, apsr80 = 42.8, apsr100 = 45.3,
                sldml20 = 36.7, sldml40 = 42.4, sldml60 = 49.0, sldml80 = 46.4, sldml100 = 50.9,
                simitate_22_4=100.0, simitate_18_4=98.8, simitate_14_4=99.4, simitate_10_4=93.1, 
                simitate_18_8 = 76.7, simitate_14_12=61.2, simitate_10_16=56.6,
                utdmhad_inter_joint_wrist_leg=80.4, utdmhad_inter_joint_leg_wrist=28.3,
                utdmhad_inter_joint_wrist_leg_eq_class_dist=18.8, utdmhad_inter_joint_leg_wrist_eq_class_dist=59.7,
                utdmhad_inter_sensor_skeleton_inertial_signals=35.5,
                utdmhad_inter_sensor_inertial_skeleton_signals=40.5,
                utdmhad_inter_sensor_skeleton_inertial_sldml=23.1,
                utdmhad_inter_sensor_inertial_skeleton_sldml=40.5,
                utdmhad_23_4_skeleton=92.7,utdmhad_19_8_skeleton=74.8,utdmhad_15_12_skeleton=81.1,utdmhad_11_16_skeleton=77.7,utdmhad_7_20_skeleton=57.2,utdmhad_3_24_skeleton=59.4,
                utdmhad_23_4_inertial=81.3,utdmhad_19_8_inertial=74.0,utdmhad_15_12_inertial=63.5,utdmhad_11_16_inertial=43.3,utdmhad_7_20_inertial=41.3,utdmhad_3_24_inertial=29.2,
                utdmhad_23_4_fused=90.2,utdmhad_19_8_fused=76.0,utdmhad_15_12_fused=78.7,utdmhad_11_16_fused=69.4,utdmhad_7_20_fused=65.0,utdmhad_3_24_fused=44.7,
                }
\usepackage[detect-none]{siunitx}
\pgfplotsset{width=.28\linewidth,compat=1.9}

\begin{document}
% \title{Multimodal One-Shot Action Recognition by Learning Signal Level Action Embeddings}

\title{SL-DML: Signal Level Deep Metric Learning for Multimodal One-Shot Action Recognition}

% \title{Using Triplet Loss for Multimodal One-Shot Action Recognition}

\author{\IEEEauthorblockN{Raphael Memmesheimer, Nick Theisen, Dietrich Paulus}
\IEEEauthorblockA{Active Vision Group\\ 
% Active Vision Group\\
University of Koblenz-Landau\\
% Atlanta, Georgia 30332--0250\\
\texttt{\{raphael, nicktheisen, paulus\}@uni-koblenz.de}}
% \and
% \IEEEauthorblockN{Homer Simpson}
% \IEEEauthorblockA{Twentieth Century Fox\\
% Email: homer@thesimpsons.com}
% \and
% \IEEEauthorblockN{James Kirk\\ and Montgomery Scott}
% \IEEEauthorblockA{Starfleet Academy\\
% San Francisco, California 96678--2391\\
% Telephone: (800) 555--1212\\
% Fax: (888) 555--1212}
}

\maketitle

\begin{abstract}
Recognizing an activity with a single reference sample using metric learning approaches is a promising research field. The majority of few-shot methods focus on object recognition or face-identification. 
We propose a metric learning approach to reduce the action recognition problem to a nearest neighbor search in embedding space. 
We encode signals into images and extract features using a deep residual CNN. Using triplet loss, we learn a feature embedding. The resulting encoder transforms features into an embedding space in which closer distances encode similar actions while higher distances encode different actions.
Our approach is based on a signal level formulation and remains flexible across a variety of modalities. It further outperforms the baseline on the large scale NTU RGB+D 120 dataset for the one-shot action recognition protocol by \ntuoneshotimpro\%. With just 60\% of the training data, our approach still outperforms the baseline approach by \ntuoneshotimproreduced\%. With 40\% of the training data, our approach performs comparably well to the second follow up.
Further, we show that our approach generalizes well in experiments on the UTD-MHAD dataset for inertial, skeleton and fused data and the Simitate dataset for motion capturing data. 
Furthermore, our inter-joint and inter-sensor experiments suggest good capabilities on previously unseen setups.
\end{abstract}


% \fi
\IEEEpeerreviewmaketitle

\section{Introduction}
Learning to identify unseen classes from a few samples is an active research topic.  Metric learning in computer vision research mainly concentrates on one-shot object recognition \cite{fu2015zero}, person re-identification \cite{yi2014deep,wojke2018deep} or face identification \cite{schroff2015facenet}. Only recently few-shot methods for action recognition \cite{careaga2019metric, jasani2019skeleton, liu2019ntu} have gained popularity. These approaches presented good results for one-shot action recognition but only concentrate on single modalities like image- or skeleton sequences. We propose to use a signal level representation that allows flexible encoding of signals into an image and the fusion of signals from different senor modalities.
% Signals can be either encoded per modality or used for the fusion of multiple modalities.
% Common methods are metric learning \cite{wojke2018deep,kliper2011one,schroff2015facenet} and meta-learning \cite{gui2018few}.

\begin{figure}
    \centering
    \includegraphics[width=\linewidth]{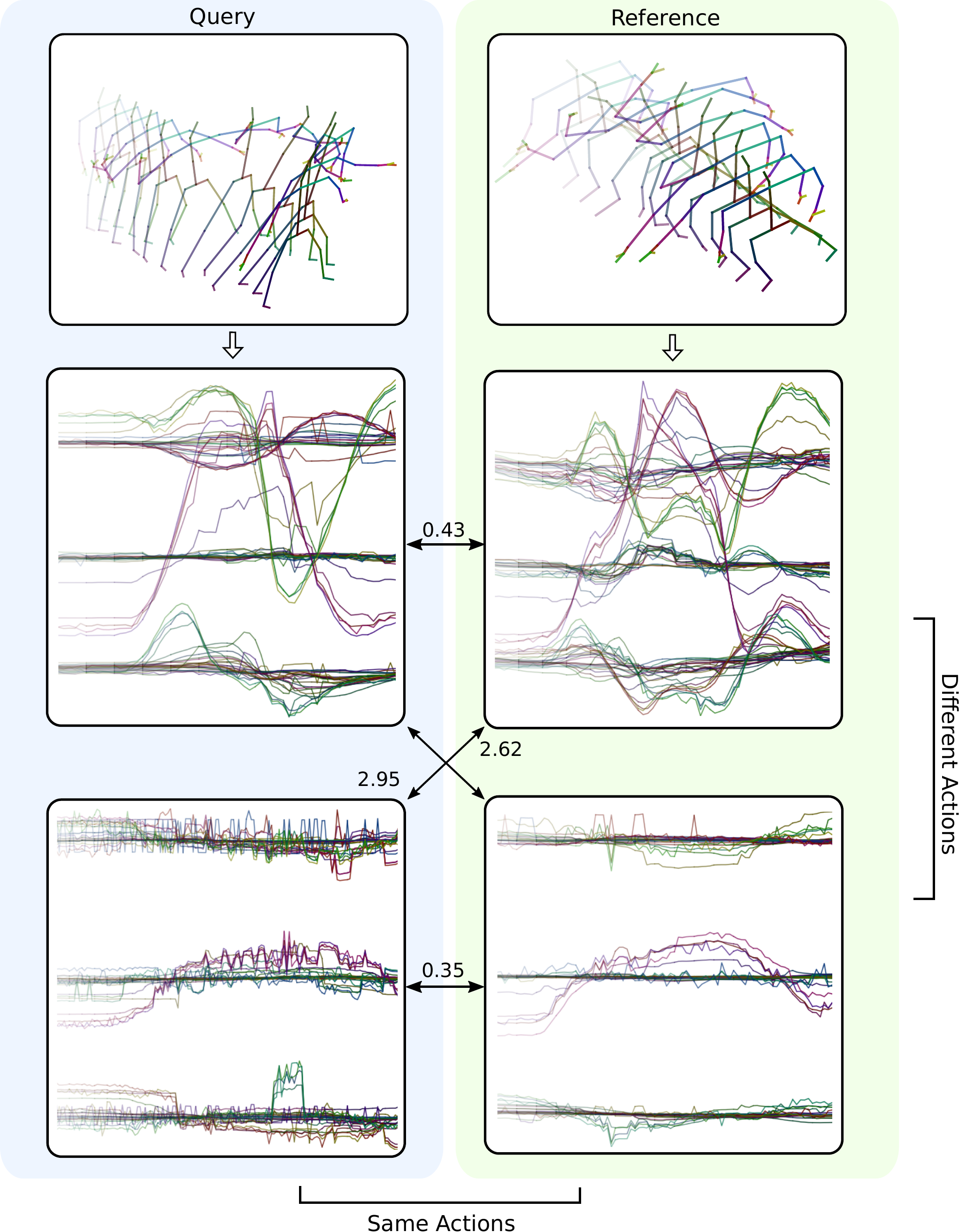}
    \caption{Illustrative example. In this example, a skeleton is transformed into an image-like representation. Joint axes are encoded as signals. Each axis in a different color. Our approach encodes an action sequence representation into an embedding vector. Low Euclidean distances on the action embedding represent close similarity, whereas higher distances represent different actions. This approach allows for ine-shot action classification or clustering similar activities. The underlying signal level representation enables multimodal applications.}
    \label{fig:overview}
\end{figure}

In contrast to classification methods, which predict class labels, metric learning approaches learn an embedding function.
% In image ranking, for instance, similar images have a lower distance in embedding space, whereas more dissimilar images have a higher distance.
% Our approach learns an embedding function that transforms a signal-image of an action into an embedding space. 
Our approach learns a function that embeds signal-images into an embedding space.
One-shot action recognition then becomes a nearest neighbor search in embedding space. \fignamelong \ref{fig:overview} gives an application example for one-shot action recognition on skeleton sequences using our approach. 
% It is to note that the signal-level image representation shown in this Figure serves for illustrative purposes. The actual proposed representation is shown below in \figname \ref{fig:representation_example}.

While it may appear implausible, initially, to encode signals into an image representation for action recognition, it has some benefits.
First, it allows generalization across different sensor modalities as long as a sensor originates multivariate signal sequences or higher-level features such as human pose estimates. There is no need for modality-specific architectures or pipelines.
Further, an image-like representation allows the usage of well-studied and well-performing classification architectures \cite{he2016deep}. 
Finally, experiments for multimodal or inter-modal one-shot action recognition can be conducted flexibly.

In our study, signals originate from 3D skeleton sequences gathered by an RGB-D camera, inertial, or motion capturing measurements. 
To fuse multiple modalities, e.g. skeleton sequences and inertial measurements, the signal matrices are concatenated and represented as an image.
Inter-modal experiments are especially interesting, as they allow training on one modality and recognition on another, previously unseen, modality by providing only a single reference. A new sensor can be used for action recognition without any prior training data from that sensor.

The main contribution of this paper is a novel model for one-shot action recognition on a signal level. A classifier and embedding encoder are jointly optimized using \textit{triplet margin loss} \cite{weinberger2009distance} in conjunction with a \textit{Muti-Similarity Miner} \cite{wang2019multi}. The nearest neighbor in embedding space defines the most similar action.
Our proposed approach lifts the state-of-the-art in one-shot action recognition on skeleton sequences on the NTU RGB+D 120 dataset for the one-shot evaluation protocol by \ntuoneshotimpro\%.
We claim that our approach based on triplet margin loss and a common signal-level representation yields high flexibility for applications in one-shot action recognition. We achieve good results on one-shot action recognition for conventional sensor modalities (skeleton sequences, inertial measurements, motion capturing measurements). Our approach shows good capabilities when being trained on one modality and inferred on a different modality by providing a single reference sample per action class of the unknown modality. This allows e.g. training on skeleton sequences and inference on inertial measurements. We provide the source code for reproduction and validation\footnote{\url{https://github.com/raphaelmemmesheimer/sl-dml}}.
%Also, different setups, e.g., training on inertial sensors placed on the wrist and application with a placement on the leg, are possible.

\section{Related Work}
We give a brief overview of methods related to metric learning and few-show recognition approaches in general. We focus on methods for action embeddings and few-shot action recognition.

% \paragraph{Action Recognition}
Our approach builds on image representation of sensor sequences. Prior work has already presented representations for action recognition with 
skeleton sequences. Wang \andothers \cite{wang2018action} encode joint trajectory maps into images based on three spatial perspectives. %\todo{spatial? visual? emotional?}
Caetano \andothers \cite{caetano2019skelemotion, caetano2019skeleton} represent a combination of reference joints and a tree-structured skeleton as images. Their approach preserves spatio-temporal relations and joint relevance. 
% Kim \andothers \cite{kim2017interpretable}.
Liu \andothers \cite{liu2017enhanced} presented a combination of skeleton visualization methods and jointly trained them on multiple streams. 
% encode signals in a slightly different way by representing the individual axes color encoded. 
In contrast to our approach, their underlying representation enforces custom network architectures and is constrained to skeleton sequences, whereas our approach adds flexibility to other sensor modalities.
% Kim \andothers \cite{kim2017interpretable} presented an interpretable visual method for action recognition using temporal convolutional networks. Their approach uses a spatio-temporal representation, which allows visual analysis to understand why a model predicted an action. Especially joint contributions are visually interpretable.

Metric learning has been intensively studied in computer vision. A focus is on metric learning from photos or cropped detection boxes for person re-identification or image-ranking.
Schroff \andothers \cite{schroff2015facenet} presented a joint face recognition and clustering approach. They trained a network such that the squared L2 distances in the embedding space directly correspond to face similarity \cite{schroff2015facenet}. Triplet loss \cite{weinberger2009distance} is used for training the embedder. The embedding minimizes distances between anchor images and positive images (i.e., same person, different viewpoint) and maximizes distances to negative samples (different person).
Yi \andothers \cite{yi2014deep} presented a deep metric learning approach based on a siamese deep neural network for person re-identification.  The two sub-nets are combined using a cosine layer.
Wojke \andothers \cite{wojke2018deep} propose a deep cosine metric learning approach for the person re-identification task. The \textit{Cosine Softmax Classifier} pushes class samples towards a defined class mean and therefore allows similarity estimation by a nearest neighbor search. 
% Triplet loss
% Magnet loss

% \paragraph{Action embedding}

A recent action embedding approach by Hahn \andothers \cite{hahn2019action2vec} takes inspiration from the success of word embeddings in natural language processing. They combine linguistic cues from class labels with spatio-temporal features from sequences. A hierarchical recurrent neural network trains a  feature extractor. 
A joint loss combines classification accuracy and similarity trains a function to encode the input into an embedding. Discriminative embeddings are important for few-shot learning approaches. Jasani \andothers \cite{jasani2019skeleton} proposed a similar approach for skeleton-based zero-shot action recognition. 
A \textit{Spatio Temporal Graph Convolution Network (ST-GCN)} \cite{yan2018spatial}  extracts features, which are encoded in semantic space by a continuous bag of words method.

% \paragraph{Few-Shot approaches}

% \paragraph{One-Shot Action Recognition}
% Kliper-Gross is hard to read. What features are used? 
One-shot action recognition is in comparison to image ranking, or person re-identification a quite underrepresented research domain.
Kliper-Gross \andothers \cite{kliper2011one} proposed One-Shot-Similarity Metric Learning. A projection matrix that improves the One-Shot-Similarity relation between the example same and not-same training pairs represents a reduced feature space \cite{kliper2011one}. 
Fanello \andothers \cite{fanello2013one} use \textit{Histogram of Flow} and \textit{Global Histogram of Oriented Gradient} descriptors with adaptive sparse coding and are classified using a linear SVM.
Careaga \andothers \cite{careaga2019metric} propose a two-stream model for few-show action recognition on image sequences. They aggregate features from optical flow and the image sequences separately by a \textit{Long Short Term Memory (LSTM)} and fuse them afterward for learning metrics.
% \cite{escalante2017principal}
Rodriguez \andothers \cite{rodriguez2017fast} presented an one-shot approach based on \textit{Simplex Hidden Markov Models (SHMM)}. Improved dense trajectories are used as base features \cite{wang2013action}. A maximum a posteriori (MAP) adoption and an optimized Expectation Maximisation reduce the feature space. A maximum likelihood classification, in combination with the SHMM, allows one-shot classification.
Roy \andothers \cite{roy2018action} propose a Siamese network approach for discriminating actions by a contrastive loss on a low dimensional representation gathered factory analysis.
Mishra \andothers \cite{mishra2018generative} presented a generative framework for zero- and few-shot action recognition on image sequences. A probability distribution models classes of actions. The parameters are functions for semantic attribute vectors that represent the action classes. 

\begin{figure*}[ht]
    \centering
    \includegraphics[width=.83\linewidth]{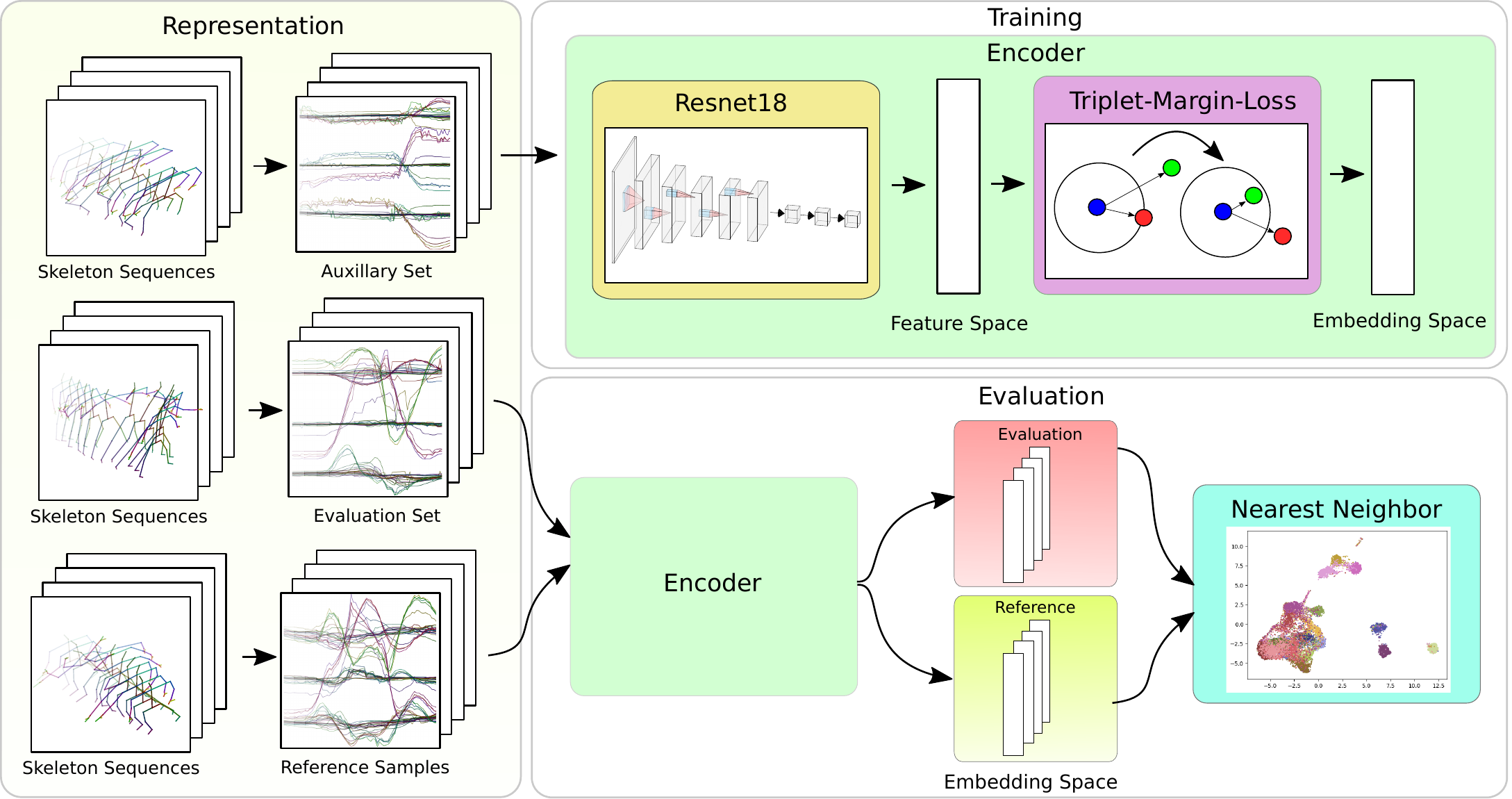}
    \caption{Approach overview: We represent actions on a signal level. In the example, we transformed skeleton joint axes into images. We use a Resnet18 architecture in conjunction with a triplet loss to train a model that transforms an image into an embedding space. For inference, the trained encoder encodes a set of references and queries. The closest reference in embedding space  represents the most similar activities for which we use a nearest-neighbor search.}
    \label{fig:approach}
\end{figure*}

Liu \andothers \cite{liu2019ntu} presented, along with the NTU RGB+D 120 dataset, an approach for one-shot action recognition. They propose an approach named Action-Part Semantic-Relevance aware (APSR). Features are generated by a ST-LSTM \cite{liu2017skeletonlstm}. Motivated by word embedding methods, Liu \andothers propose to estimate semantic relevance on body parts and the actions. Similar semantic relevance for the body parts assigns new action instances.

The field of multi-modal few-shot action recognition is entirely unexplored. Somehow related is the work of Al-Naser \andothers \cite{al2018hierarchical} who presented a zero-shot action recognition approach by combining gaze guided object recognition with a gesture recognition arm-band. Actions are detected by fusing features of sub-networks per modality and integrating action definitions. Only three actions demonstrate the recognition results.

\section{Approach}
To cover the action recognition task across a variety of sensor modalities we consider the action recognition problem on a signal level. Signals are encoded in a discriminate image representation. An image-like representation allows direct adaption of already established image classification architectures for extracting features. On the extracted features we train a similarity function yielding an action embedding using triplet loss. The triplet loss minimizes embedding distances between similar action samples while maximizing distances between different actions. Finally to solve the one-shot problem, we apply a nearest neighbor search in the embedding space. An illustration of our approach is given in \figname \ref{fig:approach}.

\subsection{Problem Formulation}

We consider the one-shot action recognition problem as a metric learning problem. First we encode action sequences on a signal level into an image representation.
The input in our case is a signal matrix $\Signals \in \SignalSpace$ where each row vector represents a discrete 1-dimensional signal and each column vector represents a sample of all sensors at one specific time step. The matrix is transformed to an RGB image $\Image \in \ImageSpace$ by normalizing the signal length $M$ to $W$ and the range of the signals to $H$. The identity of each signal is encoded in the color channel.
This results in a dataset $\Dataset = \{(\Image_i, \Label_i)\}_{i=1}^{\NumData}$ of~$\NumData$ training images $\Image_{1, \dots, \NumData}$ with labels $\Label_i\in\{1, \dots, \NumLabels\}$.
Our goal is to train a feature embedding $\Feature = \Encode[\EncoderParams]{\Image}$ with parameters~$\EncoderParams$ which projects input images~$\Image\in\ImageSpace$ into a feature representation~$\Feature \in \FeatureSpace$. The feature representation reflects minimal distances for \textit{similar} classes.
% \item Benefit: Metric learning formulation over a standard classification generalizes to unseen classes through the similarity function measure~$\odot$.
\subsection{Representations}    
\begin{figure}
    \centering
    \includegraphics[width=\linewidth]{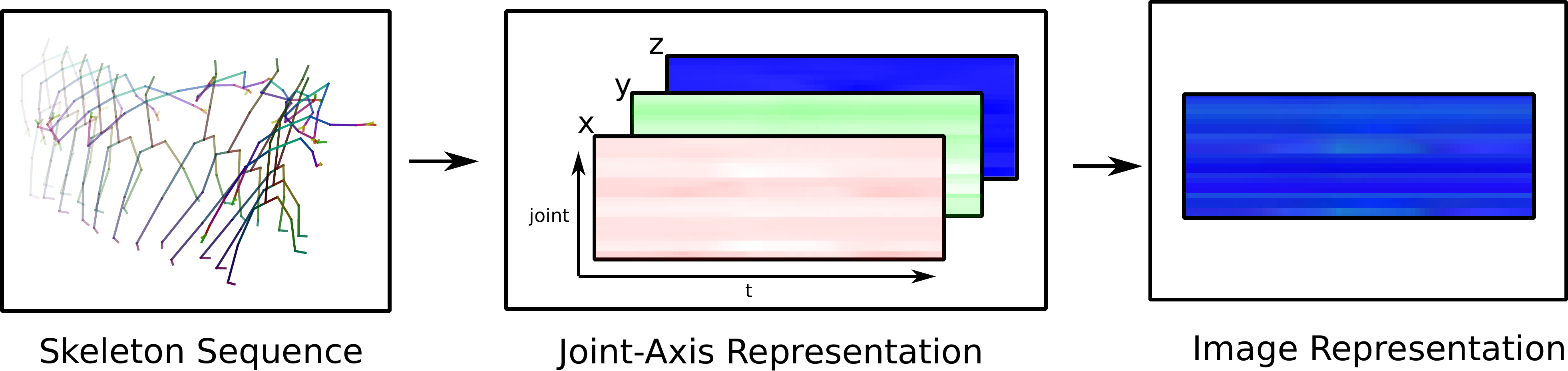}
    \caption{Examplary representation for a throwing activity of the NTU-RGB+D 120 dataset.}
    \label{fig:representation_example}
\end{figure}
% \begin{figure*}[th] \centering$
%   \vspace{0.06in}
%   \begin{array}{cccc}
%       \includegraphics[width=.2\linewidth]{images/examples/a1_s7_t2_skeleton} &
%       \includegraphics[width=.2\linewidth]{images/examples/a1_s7_t2_inertial} &
%       \includegraphics[width=.2\linewidth]{images/examples/a1_s7_t2_fused} &
%       \includegraphics[width=.2\linewidth]{images/examples/bring_glis_blue_cabinet_to_kitchen_table_2018-09-21-15-01-01.png} \\
%       \includegraphics[width=.2\linewidth]{images/examples/swap_a1_s7_t2_skeleton.mat.png} &
%       \includegraphics[width=.2\linewidth]{images/examples/swap_a1_s7_t2_inertial.mat.png} &
%       \includegraphics[width=.2\linewidth]{images/examples/swap_a1_s7_t2_fused} &
%       \includegraphics[width=.2\linewidth]{images/examples/swap_bring_glis_blue_cabinet_to_kitchen_table_2018-09-21-15-01-01.png} \\
%       (a) & (b) & (c) & (d)
%   \end{array}$
%   \caption{Example representations for skeleton sequences \textit{(a)}, inertial measurements  \textit{(b)}, fused measurements \textit{(c)} and motion capturing data  \textit{(d)}. The three example representations show the range of modalities we conducted one-shot experiments on. One highlight is that our approach allows training on skeleton data like in \textit{(a)} and recognize actions from inertial data \textit{(b)} with a single reference sample of the previously unseen modality.}
%   \label{fig:representation_examples}
% \end{figure*}
Our approach builds upon a discriminable image representation. Therefore, we propose a novel, compact signal level representation. Multivariate signal or higher-level feature sequences are reassembled into a 3 channel image. Each row of the resulting image corresponds to one joint and each channel corresponds to one sample in the sequence. The color channels, red, green and blue, represent respectively the signals' x-, y- and z-values. The resulting images are normalized to the range of $\numrange{0}{1}$. We chose to normalize over the whole image to preserve the relative magnitude of the signals. In contrast to the representations used for multimodal action classification \cite{memmesheimer2020gimme} or skeleton based action recognition \cite{wang2018action,liu2017enhanced} the proposed representation is invertible and more compact. The construction of the representation is depicted in \figname \ref{fig:representation_example}.

\subsection{Feature Extraction}

Most action recognition approaches based on CNNs present custom architecture designs in their pipelines \cite{liu2017enhanced}. A benefit is the direct control over the number of model parameters that can be specifically engineered for data representations or use cases. Recent advances in architecture design cannot be transferred directly. Searching good hyper-parameters for training is then often an empirical study. Minor architecture changes can result in a completely different set of hyper-parameters.
He \andothers \cite{he2016deep} suggested the use of residual layers during training to tackle the vanishing gradient problem. We take advantage of the recent development in architecture design and decided to use a Resnet18 \cite{he2016deep} architecture. For weight initialization we use a pre-trained model. 
% \textcolor{red}{MOre argumentation and description, may rewrite, as current text is based on gimme signals}. 
After the last feature layer we use a two-layer perceptron to transform the features into the embedding size. The embedding is refined by the metric learning approach.

\subsection{Metric Learning}
\label{ssec:metric_learning}

Metric learning aims to learn a function to transform an image into an embedding space, where the embedding vectors of similar samples are encouraged to be closer, while dissimilar ones are pushed apart from each other \cite{wang2019multi}. We use a triplet loss in combination with a \textit{Multi-Similarity-Miner} \cite{wang2019multi} for mining good triplet candidates during training. 

% \begin{itemize}
%     \item Triplet margin loss
%     \item Angular loss
%     \item Describe margins, embedding size ...
% \end{itemize}

% \subsubsection{Triplet Loss}

While the triplet loss has been used in image ranking \cite{bui2017compact}, face recognition \cite{schroff2015facenet}, person re-identification \cite{hermans2017defense} it has only 
rarely been used for inter- and cross-modal ranking to improve action recognition \cite{wang2018cooperative} and for complex event detection \cite{hou2017content}.
Given a triplet of an anchor image $\Image_\circ$, a positive data sample, representing the same action class image $\Image_\uparrow$ and a negative sample, representing a different action class $\Image_\downarrow$ the triplet loss can be formulated as:

\begin{equation}
    \begin{aligned}
    % This kind of reads shitty in tex, but with multi aligned equations you can get rid of compilation errors using the \left. and \right.
    \LossVar_{t}\left(\Image_\circ, \Image_\uparrow, \Image_\downarrow\right)= 
    \max\left( \right.
          &\|\Encode{\Image_\circ}-\Encode{\Image_\uparrow}\|_{2} -\\
          & {\|\Encode{\Image_\circ} -\Encode{\Image_\downarrow}\|}_{2}+ \\
          & \delta ,0\left.\right),
    \end{aligned}
\end{equation}
where $\delta$ describes an additional distance margin.

Finding good candidate pairs is crucial. Therefore we use a \textit{Multi-Similarity Miner} \cite{wang2019multi, Musgrave2019} to mine positive and negative pairs that are assumed to be difficult to push apart in the embedding space.
That means positive pairs are constructed by an anchor and positive image pair $\{\Image_\circ,\Image_\uparrow\}$ and its embedding $\Encode{\Image_\circ}$, preferring pairs with a high distance in embedding space with the following condition:
\begin{equation}
    \label{eq-select-pos}
    \|\Encode{\Image_\circ}-\Encode{\Image_\uparrow}\|_{2} > \min_{\Label_k \neq \Label_i} \|\Encode{\Image_i}-\Encode{\Image_k}\|_{2} - \epsilon,
\end{equation}
likewise, negative pairs $\{\Image_\circ,\Image_\downarrow\}$ are mined by the lowest distance in embedding space:
\begin{equation}
    \label{eq-select-neg}
    \|\Encode{\Image_\circ}-\Encode{\Image_\downarrow}\|_{2} < \max_{\Label_k \neq \Label_i} \|\Encode{\Image_i}-\Encode{\Image_k}\|_{2} + \epsilon,
\end{equation}
where $\epsilon$ is a given margin.
% \textcolor{red}{May simplify the formulation of the similarity calculation.}
Finally, we yield the total loss by:

\begin{equation}
    \LossVar = \alpha \LossVar_{t} + \beta \LossVar_{c},
\end{equation}
such that the influences of the loss can be weighted using the scalars $\alpha$ for the triplet loss $\LossVar_{t}$ and $\beta$ for the classifier loss $\LossVar_{c}$. We utilize a cross entropy loss for $\LossVar_{c}$. Finding an action class by a query and set of references is now reduced to a nearest-neighbor search in the embedding space. 
The classifier and encoder are jointly optimized. After the last feature layer of the classifier a two-layer perceptron is used to to yield an embedding size of 128.
%using a linear layer. 
% \textcolor{red}{Check if the weightning is correct.}
% \subsection{Nearest neighbor Search}

% \textcolor{red}{\lipsum[1]}

% \subsection{Encoder}

\subsection{Implementation}

Our implementation is based on PyTorch \cite{Musgrave2019}, \cite{paszke2019pytorch}. 
% The baseline representations were generated as described in \cite{memmesheimer2020gimme}. 
We tried to avoid many of the metric learning flaws as pointed out by Musgrave \andothers \cite{musgrave2020metric} by using their training setup and hyperparameters where applicable.
% and therefore use their training setup and set of hyperparameters as far as possible for better comparison.
Key differences are that we use a Resnet18 \cite{he2016deep} architecture and avoid the proposed four-fold cross validation for hyperparameter search in favour of better comparability to the proposed one-shot protocol on the NTU-RGB+D dataset \cite{liu2019ntu}. Note, we did not perform any optimization of the hyperparameter.
A batch size of 32 was used on a single Nvidia GeForce RTX 2080 TI with 11GB GDDR-6 memory. We trained for 100 epochs with initialized weights of a pre-trained Resnet18 \cite{he2016deep}. 
% Embeddings and models were saved for the epoch with the highest accuracy. 
The classification and metric loss were weighted by $0.5$ unless stated otherwise. For the multi similarity miner we used an epsilon of 
$0.05$ 
while we used a margin of $0.1$ for the triplet margin loss. A 
% stochastic gradient decent 
RMSProp optimizer
with a learning rate of 
% $0.01$
$10^{-6}$
was used in all optimizers. The embedding model outputs a 128 dimensional embedding and the classifier yielded a 128 dimensional feature vector.
% \textcolor{red}{\lipsum[1]}

\section{Experiments}

\begin{figure*}[t] \centering$
  \vspace{0.06in}
  \begin{array}{ccc}
        % \begin{figure}[t]
% %   \vspace{0.06in}
% %   \begin{array}{ccc}
%   %TBM1
      \begin{tikzpicture}
            \begin{axis}[
                xlabel={\#Training Classes},
                ylabel={Accuracy},
                xmin=20, xmax=100,
                ymin=25, ymax=55,
                xtick={20,40,60,80,100},
                ytick={0,10,20,25,30,35,40,45,50,55},
                legend pos=south east,
                legend style={nodes={scale=0.6, transform shape}},
                ymajorgrids=true,
                grid style=dashed,
            ]
            \addplot[color=blue, mark=square] coordinates {(20,\results{sldml20})(40,\results{sldml40})(60,\results{sldml60})(80,\results{sldml80})(100,\results{sldml100})};\addlegendentry{Ours}
            \addplot[color=red, mark=square] coordinates {(20,\results{apsr20})(40,\results{apsr40})(60,\results{apsr60})(80,\results{apsr80})(100,\results{apsr100})};\addlegendentry{APSR \cite{liu2019ntu}}
            \end{axis}
        \end{tikzpicture}
%   \caption{Results for different auxiliary set sizes.}
%   \label{fig:resultsauxilllary}
% \end{figure}
        &
        
        % \begin{figure}[t]
% %   \vspace{0.06in}
% %   \begin{array}{ccc}
%   %TBM1
      \begin{tikzpicture}
            \begin{axis}[
                xlabel={\#Training Classes},
                ylabel={Accuracy},
                xmin=3, xmax=23,
                ymin=20, ymax=100,
                xtick={3,7,11,15,19,23},
                ytick={0,10,20,30,40,50,60,70,80,90},
                legend pos=south east,
                legend style={nodes={scale=0.5, transform shape}},
                ymajorgrids=true,
                grid style=dashed,
            ]
            % \addplot[color=red, mark=square] coordinates {(3, \utdmhadresultskeletonauxthreevalfour)(7, \utdmhadresultskeletonauxsevenvalfour)(11, \utdmhadresultskeletonauxelevenvalfour)(15, \utdmhadresultskeletonauxfifteenvalfour)(19,\utdmhadresultskeletonauxnineteenvalfour)(23,\utdmhadresultskeletonauxtwentythreevalfour)};\addlegendentry{Skl s. val}
            
            % \addplot[color=black, mark=square] coordinates {(3, \utdmhadresultimuauxthreevalfour)(7, \utdmhadresultimuauxsevenvalfour)(11, \utdmhadresultimuauxelevenvalfour)(15, \utdmhadresultimuauxfifteenvalfour)(19,\utdmhadresultimuauxnineteenvalfour)(23,\utdmhadresultimuauxtwentythreevalfour)};\addlegendentry{IMU s. val}
            
            % \addplot[color=magenta, mark=square] coordinates {(3, \utdmhadresultfusedauxthreevalfour)(7, \utdmhadresultfusedauxsevenvalfour)(11, \utdmhadresultfusedauxelevenvalfour)(15, \utdmhadresultfusedauxfifteenvalfour)(19,\utdmhadresultfusedauxnineteenvalfour)(23,\utdmhadresultfusedauxtwentythreevalfour)};\addlegendentry{Fused s. val}
            
            \addplot[color=green, mark=square] coordinates {
                (3,  \results{utdmhad_3_24_skeleton})
                (7,  \results{utdmhad_7_20_skeleton})
                (11, \results{utdmhad_11_16_skeleton})
                (15, \results{utdmhad_15_12_skeleton})
                (19, \results{utdmhad_19_8_skeleton})
                (23, \results{utdmhad_23_4_skeleton})
            };
            \addlegendentry{Skl}

            \addplot[color=blue, mark=square] coordinates {
                (3,  \results{utdmhad_3_24_inertial})
                (7,  \results{utdmhad_7_20_inertial})
                (11, \results{utdmhad_11_16_inertial})
                (15, \results{utdmhad_15_12_inertial})
                (19, \results{utdmhad_19_8_inertial})
                (23, \results{utdmhad_23_4_inertial})
            };
            \addlegendentry{IMU}
    
            \addplot[color=red, mark=square] coordinates {
                (3,  \results{utdmhad_3_24_fused})
                (7,  \results{utdmhad_7_20_fused})
                (11, \results{utdmhad_11_16_fused})
                (15, \results{utdmhad_15_12_fused})
                (19, \results{utdmhad_19_8_fused})
            };\addlegendentry{Fused}

            \end{axis}
        \end{tikzpicture}
%   \caption{Results for different auxiliary set sizes on the UTD-MHAD dataset.}
%   \label{fig:resultsutdmhad}
% \end{figure}
        & 
        \begin{tikzpicture}%[scale=0.7]
    \begin{axis}[
        xlabel={\#Training Classes},
        ylabel={Accuracy},
        xmin=10, xmax=22,
        ymin=50, ymax=100,
        xtick={10,14,18,22},
        ytick={0,10,20,30,40,50,60,70,80,90},
        legend pos=south east,
        legend style={nodes={scale=0.6, transform shape}},
        ymajorgrids=true,
        grid style=dashed,
    ]
    \addplot[color=red, mark=square] coordinates {(22,\results{simitate_22_4})(18,\results{simitate_18_8})(14,\results{simitate_14_12})(10,\results{simitate_10_16})};\addlegendentry{Ours}
    \addplot[color=blue, mark=square] coordinates {(22,\results{simitate_22_4})(18,\results{simitate_18_4})(14,\results{simitate_14_4})(10,\results{simitate_10_4})};\addlegendentry{Ours s. val}
    \end{axis}
\end{tikzpicture}\\
        \textit{(a)} & \textit{(b)} & \textit{(c)}
  \end{array}$
  \caption{Result graphs for the NTU RGB+D 120 dataset \textit{(a)}, the UTD-MHAD dataset \textit{(b)} and the Simitate dataset \textit{(c)}. \textit{Skl} denotes skeleton data, \textit{IMU} denotes inertial data, \textit{Fused} denotes multimodal data consisting of inertial- and skeleton data \textit{s. val} denotes a static evaluation set.
  }
  \label{fig:result_figs}
\end{figure*}
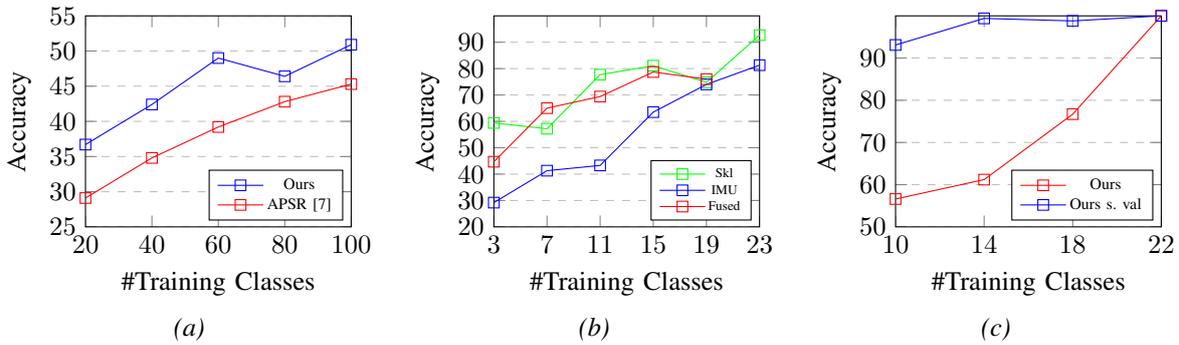

To show the multi-modal one-shot recognition performance we applied our methods to three datasets containing three different modalities. 
We used skeleton sequences from the NTU RGB+D 120 \cite{liu2019ntu} dataset for large scale one-shot action recognition. With 100 auxiliary classes and 20 evaluation classes it is the largest dataset that we applied to our approach. To show the multi modal capabilities of our approach we also used the UTD-MHAD \cite{chen2015utd} dataset (inertial and skeleton data) and the Simitate \cite{Memmesheimer2019SAH} dataset (motion capturing data).

The datasets are split into an auxiliary set, representing action classes that are used for training, and an evaluation set. In our experiments the evaluation set does contain novel actions or actions from a novel sensor modality. One sample of each test class serves as reference demonstration. This protocol is based on the one proposed by \cite{liu2019ntu} for the NTU RGB+D 120 dataset. We conducted similar experiments with the remaining two data sets. In depth descriptions are given below in \secname \ref{ssec:datasets} per dataset. Results are discussed after the dataset presentation in \secname \ref{ssec:results}.
First we trained a model on the auxiliary set. The resulting model estimates embeddings for the reference actions and then for the evaluation actions. We then calculate the nearest neighbour from the evaluation embeddings to the reference embeddings. This yields to which action from the reference set the current evaluation sample comes closest.

\subsection{Datasets}
\label{ssec:datasets}

\begin{table}[tb]
    \caption{One-shot action recognition results on the NTU RGB+D 120 dataset.}
	\begin{center}
        \small
		\begin{tabular}{lr}
% 			\toprule
            Approach                                       &  Accuracy [\%]\\
            \toprule
            Attention Network \cite{liu2017global}       &  41.0                   \\ %\midrule
            Fully Connected \cite{liu2017global}         &  42.1                   \\ %\midrule
			Average Pooling \cite{liu2017skeleton}       &  42.9                   \\ %\midrule
            APSR \cite{liu2019ntu}                       &  45.3          \\  %\midrule
            % Ours                                         &  \textbf{44.7}\% (ResNet18)          \\
            % Ours                                         &  \textbf{45.77}\% (ResNet152)          \\
            % \hline 
            % Loss weighting ($\alpha = 1, \beta = 0$)                    & \ntuoneshotresulttriplett\%  \\
            % Loss weighting ($\alpha = 0, \beta = 1$)                    & \ntuoneshotresultcrossentropy\%  \\
            Ours                                         & \textbf{\ntuoneshotresult}  \\ %tiplett magin loss (margin 0.2, SDG LR=0.1)
            %INFO:root:defaultdict(<class 'dict'>, {'samples': {'epoch': 10, 'accuracy': 0.4867513037981352}, 'val': {'epoch': 10, 'accuracy': 0.4867513037981352}})
            \bottomrule
		\end{tabular}
	\end{center}

	\label{tab:oneshot_results}
\end{table}

\begin{table}[tb]
	\caption{Results for different auxiliary training set sizes for one-shot recognition on the NTU RGB+D 120 dataset.}
	\begin{center}
        \small
		\begin{tabular}{ccr}
			\toprule
			\#Train Classes   &  APSR \cite{liu2019ntu} [\%]& Ours ($\alpha, \beta = 0.5$) [\%] \\
			%\hline\hline
            \toprule
             $20$               &     \results{apsr20}   & \textbf{\results{sldml20}}   \\ 
             $40$               &     \results{apsr40}   & \textbf{\results{sldml40}}  \\ 
             $60$               &     \results{apsr60}   & \textbf{\results{sldml60}}   \\ 
             $80$               &     \results{apsr80}   & \textbf{\results{sldml80}}   \\ 
			 $100$              &     \results{apsr100}  & \textbf{\results{sldml100}}   \\ %\midrule
            \bottomrule
		\end{tabular}
	\end{center}

	\label{tab:ontshot2_results2}
\end{table}

\begin{table}

    \caption{Ablation study for our proposed one-shot action recognition approach on the NTU RGB+D 120 dataset.}
	\begin{center}
        \small
		\begin{tabular}{llrrr}
% 			\toprule
            %Representation &
            Miner & $\alpha$ & $\beta$ &  Accuracy [\%] \\
            \toprule
            % Re-indexed & & & & \\
            % Re-indexed & & & & \\
            % Re-indexed & & & & \\
            % SwapAxis & & & & \\
            % SwapAxis & & & & \\
            % SwapAxis & & & & \\
             Triplet Margin \cite{schroff2015facenet} &1.0 & 0.0         & \ntuoneshotresulttriplettripletmarginminer  \\
             Triplet Margin \cite{schroff2015facenet} &0.0 & 1.0         & \ntuoneshotresultcrossentropytripletmarginminer  \\
             Triplet Margin \cite{schroff2015facenet} &0.5 & 0.5         & \ntuoneshotresulttripletcrossentropytripletmarginminer\\
             Multi Similarity \cite{wang2019multi} &1.0 & 0.0         & \textbf{\ntuoneshotresulttriplet}  \\
             Multi Similarity \cite{wang2019multi} &0.0 & 1.0         & \ntuoneshotresultcrossentropy  \\
             Multi Similarity \cite{wang2019multi} &0.5 & 0.5                                         & \ntuoneshotresult  \\ %tiplett magin loss (margin 0.2, SDG LR=0.1)
            %INFO:root:defaultdict(<class 'dict'>, {'samples': {'epoch': 10, 'accuracy': 0.4867513037981352}, 'val': {'epoch': 10, 'accuracy': 0.4867513037981352}})
            \bottomrule
		\end{tabular}
	\end{center}

	\label{tab:oneshot_results_ntu_ablation}
\end{table}

\paragraph{NTU RGB+D 120}
The NTU RGB+D 120 \cite{liu2019ntu} dataset is a large scale action recognition dataset containing RGB+D image streams and skeleton estimates. 
The dataset consists of 114,480 sequences containing 120 action classes from 106 subjects in 155 different views.
We follow the one-shot protocol as described by the dataset authors. The dataset is split into two parts: an auxiliary set and an evaluation set. The action classes of the two parts are distinct. 100 classes are used for training, 20 classes are used for testing. The unseen classes and reference samples are documented in the accompanied dataset repository\footnote{\url{https://github.com/shahroudy/NTURGB-D}}. \textit{A1, A7, A13, A19, A25, A31, A37, A43, A49, A55, A61, A67, A73, A79, A85, A91, A97, A103, A109, A115} are previously unseen. As reference the demonstration for filenames starting with \textit{S001C003P008R001*} are used for actions with IDs below 60 and \textit{S018C003P008R001*} for actions with IDs above 60.
One-shot action recognition results are given in \tabname \ref{tab:oneshot_results}. Like Liu \andothers \cite{liu2019ntu} we also experimented with the effect of the auxiliary set reduction. Results are given in \figname \ref{fig:result_figs} \textit{(a)} and \tabname \ref{tab:ontshot2_results2}. Further we inspect the influence of different loss weighting parameters and compare two miners: Triplet Margin \cite{schroff2015facenet} and the Multi Similarity Miner \cite{wang2019multi} in \tabname \ref{tab:oneshot_results_ntu_ablation}.

\paragraph{UTD-MHAD}

The UTD-MHAD \cite{chen2015utd} contains 27 actions of 8 individuals performing 4 repetitions each. RGB-D camera, skeleton estimates and inertial measurements are included.
The RGB-D camera is placed frontal to the demonstrating person. The IMU is either attached at the wrist or the leg during the movements. No one-shot protocol is defined therefore we defined custom splits. We started with 23 auxiliary classes and evaluated with reduced training sets. %We then proceeded two-fold. 
We evaluated our approach by moving auxiliary instances over to the evaluation set. By this we decreased the training set while increasing the evaluation set.
%We evaluated with a static evaluation set of the four actions with the highest ID (24-27) and incrementally reduced the training set. Per experiment we removed the top four action classes until the evaluation classes exceeded the auxiliary classes. 
% Second we proceeded similar for the auxiliary classes but moved the reduced training instances over into the evaluation set. By this we decreased the training set while increasing the evaluation set. 
Results for the experiments executed on skeleton, inertial and fused data are given in \tabname \ref{tab:oneshot_utdmhad}. \fignamelong \ref{fig:result_figs} \textit{(b)} show visually the influence of the auxiliary set.
In a third experiment we evaluated the inter-joint one-shot learning abilities of our approach. For actions with ids up to 21 the inertial unit was placed on the subjects wrist and for the remaining ids from 22-27 the sensor was placed on the subjects leg. This allows us to inspect the one-shot action recognition transfer to other sensor positions by learning on wrist sequences and recognize on leg sequences with one reference example.
We always used the first trial of the first subject as reference sample and the remainder for testing.
Results for the inter-joint experiment on inertial data are given in \tabname \ref{tab:oneshot_utdmhadinterjoint}. For the fused experiments we concatenated
$\Signals_{\rm fused} = (\Signals_{\rm imu}|\Signals_{\rm skl})$, where $\Signals_{\rm imu}$ denotes the inertial signal matrix and $\Signals_{\rm skl}$ denotes the skeleton signal matrix. Concatenation is only possible with equal column matrices, therefore we subsampled the modality with the higher signal sample rate.

\paragraph{Simitate}

We further evaluate on the Simitate dataset. The Simitate benchmark focuses on robotic imitation learning tasks. Hand and object data are provided from a motion capturing system in 1932 sequences containing 26 classes of 4 different complexities. The individuals execute tasks of different kinds of activities from drawing motions with their hand over to object interactions and more complex activities like ironing. 
We consider one action class of each complexity level as unknown. Namely, \textit{zickzack} from \textit{basic motions}, \textit{mix} from \textit{motions}, \textit{close} from \textit{complex} and \textit{bring} from \textit{sequential}. Resulting in an auxiliary set of 22 classes and 4 evaluation classes. The corresponding first sequence by filename is used as reference sample. Results are given in \tabname \ref{tab:oneshot_simitate_results}.

\begin{figure*}[th] \centering$
  \vspace{0.06in}
  \begin{array}{ccccc}
      \includegraphics[width=.28\linewidth]{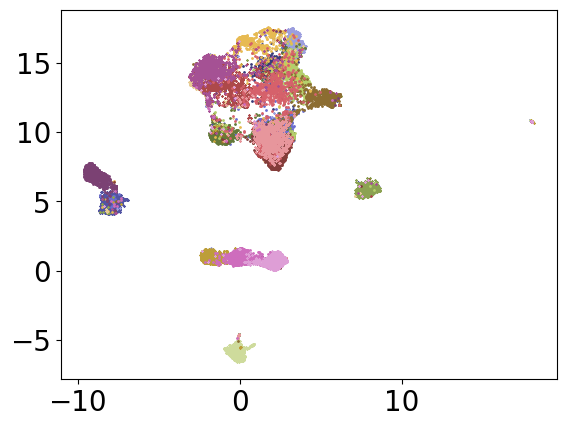} &
    %   &
      \includegraphics[width=.28\linewidth]{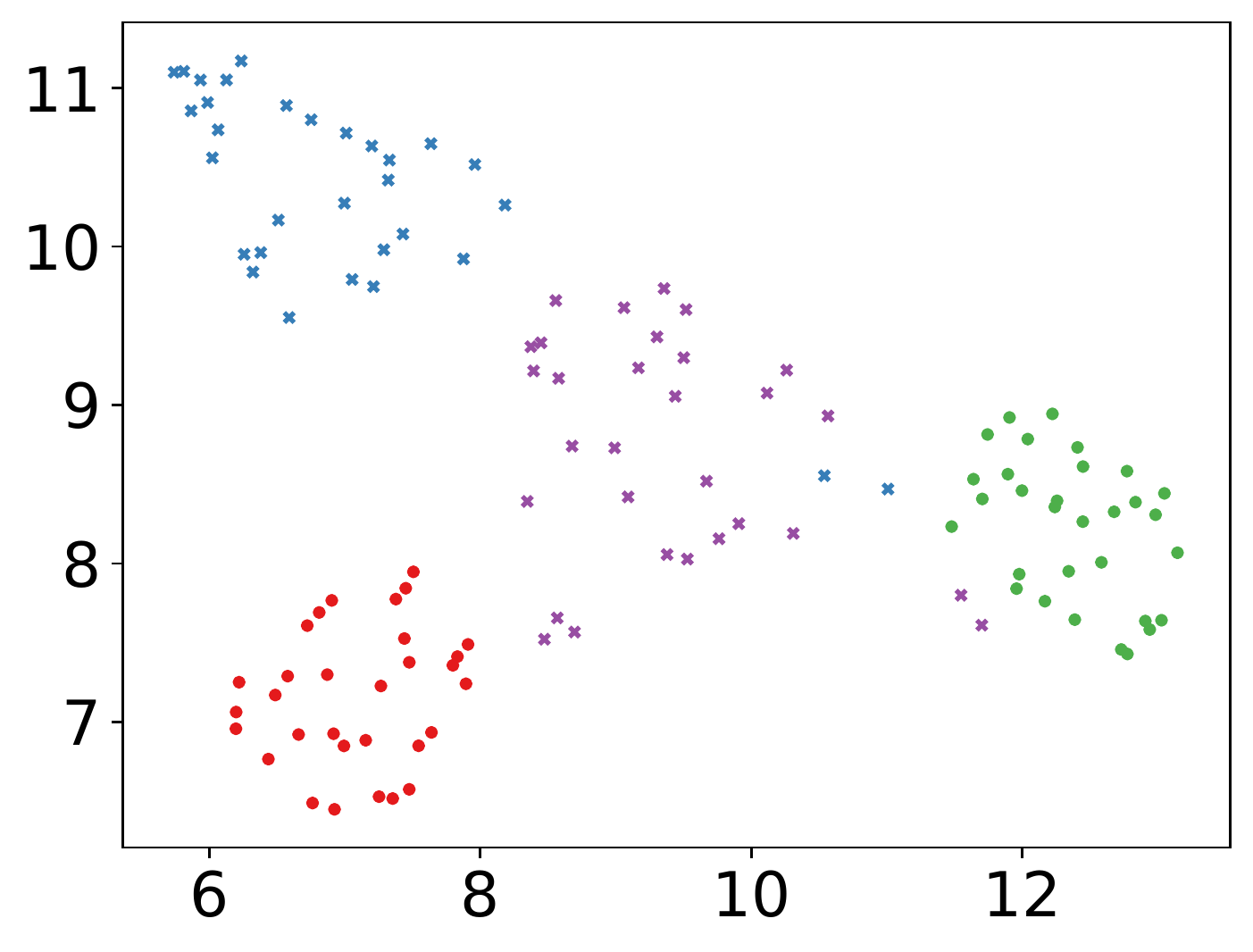} &
      \includegraphics[width=.28\linewidth]{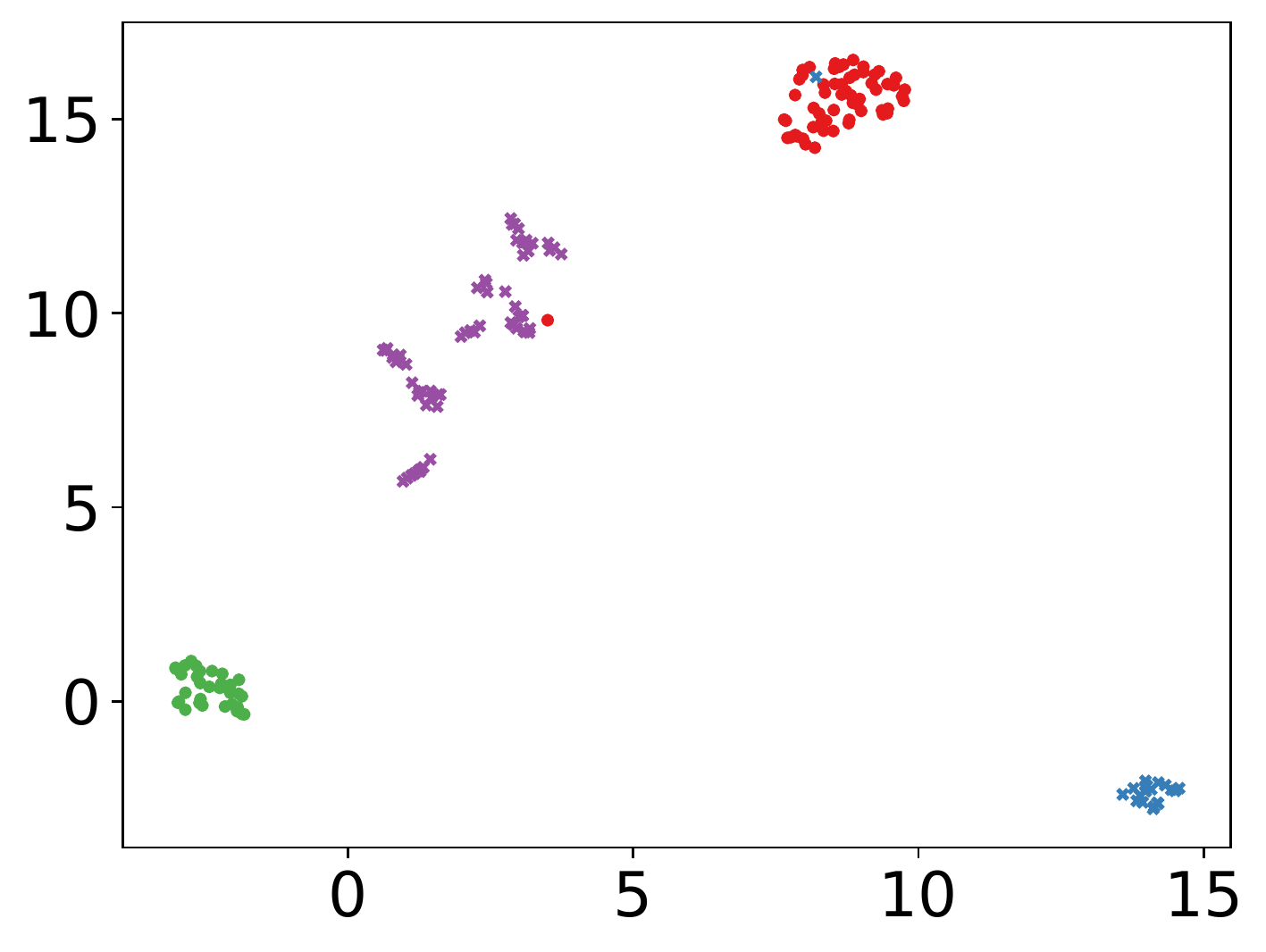}  \\
      (a) & (b) & (c)
  \end{array}$
  \caption{UMAP embedding visualization for the one-shot experiments using the NTU RGB+D 120 \textit{(a)} dataset, the UTD-MHAD dataset (IMU) \textit{(b)} and the Simitate dataset \textit{(c)}.}
  \label{fig:umap_embedding}
\end{figure*}

% \begin{figure*}[th] \centering$
%   \vspace{0.06in}
%   \begin{array}{ccccc}
%       \includegraphics[width=.2\linewidth]{images/umap_ntu} &
%       \includegraphics[width=.2\linewidth]{images/umap_utdmhad_imu_023_004} &
%       \includegraphics[width=.2\linewidth]{images/umap_simitate_aux_022_val_004} &
%       \includegraphics[width=.2\linewidth]{images/umap/umap_utdmhad_sensor_imu_skeleton.pdf} &
%       \includegraphics[width=.2\linewidth]{images/umap/umap_utdmhad_sensor_skeleton_skeleton.pdf} \\
%       (a) & (b) & (c) & (d) & (e)
%   \end{array}$
%   \caption{UMAP embedding visualization for the One-Shot experiments using the NTU RGB+D 120 \textit{(a)} dataset, the UTD-MHAD dataset (IMU) \textit{(b)} and the Simitate dataset \textit{(c)}.}
%   \label{fig:umap_embedding}
% \end{figure*}

\begin{table}[tb]
    \vspace{0.06in}
    \caption{one-shot action recognition results on the UTD-MHAD dataset.}	
	\begin{center}
        \small
		\begin{tabular}{ccrrr}
% 			\toprule
             \#Train Cl.  & \#Val Cl. &  Skl. [\%]& Inertial [\%]& Fused [\%]\\
            \toprule
            
                        %skeleton 'val': {'epoch': 19, 'accuracy': 0.967479674796748}})
            %inertial 'val': {'epoch': 3, 'accuracy': 0.8130081300813008}})
            23                   & 4       & \textbf{\results{utdmhad_23_4_skeleton}} & \results{utdmhad_23_4_inertial} & \results{utdmhad_23_4_fused} \\ \midrule%
            19                   & 8       & \results{utdmhad_19_8_skeleton} & \results{utdmhad_19_8_inertial} & \textbf{\results{utdmhad_19_8_fused}}\\ %
            %skeleton 'val': {'epoch': 7, 'accuracy': 0.7162162162162162}})
            %inertial val': {'epoch': 19, 'accuracy': 0.6675675675675675}})
            15                   & 12       & \results{utdmhad_15_12_skeleton} & \results{utdmhad_15_12_inertial} & \textbf{\results{utdmhad_15_12_fused}}\\ 
            % skeleton 'val': {'epoch': 7, 'accuracy': 0.6214574898785425}}
            % inertial 'val': {'epoch': 15, 'accuracy': 0.5504761904761905}})
            11                   & 16       & \textbf{\results{utdmhad_11_16_skeleton}} &
            \results{utdmhad_11_16_inertial} & 
            \results{utdmhad_11_16_fused}\\
            7                   & 20       & \results{utdmhad_7_20_skeleton} &
            \results{utdmhad_7_20_inertial} & \textbf{\results{utdmhad_7_20_fused}}\\
            3                   & 24       & \textbf{\results{utdmhad_3_24_skeleton}} &
            \results{utdmhad_3_24_inertial}& \results{utdmhad_3_24_fused}\\
            
            \bottomrule
		\end{tabular}
	\end{center}

	\label{tab:oneshot_utdmhad}
\end{table}

\begin{table}[tb]
    \caption{Inter-joint one-shot action recognition results on the UTD-MHAD dataset.}	
	\begin{center}
        \small
		\begin{tabular}{cclr}
% 			\toprule
             \#Train Classes  & \#Val Classes  &  Train Joint & Accuracy [\%] \\
            \toprule
            21 & 6       &  Left wrist & \results{utdmhad_inter_joint_wrist_leg}   \\
            6  & 21      &  Left leg & \results{utdmhad_inter_joint_leg_wrist} \\ \midrule
            6  & 6       &  Left wrist & \results{utdmhad_inter_joint_wrist_leg_eq_class_dist}   \\
            6  & 6       &  Left leg & \results{utdmhad_inter_joint_leg_wrist_eq_class_dist} \\
            \bottomrule
		\end{tabular}
	\end{center}

	\label{tab:oneshot_utdmhadinterjoint}
\end{table}

\begin{table}[tb]
    \caption{Inter-sensor one-shot action recognition results on the UTD-MHAD dataset.}	
	\begin{center}
        \small
		\begin{tabular}{cccr}
% 			\toprule
             Train Modality & Val. Modality & Representation & Accuracy [\%]\\
            \toprule
           Skeleton & Inertial & Signals \cite{memmesheimer2020gimme} & \results{utdmhad_inter_sensor_skeleton_inertial_signals}   \\
           Inertial & Skeleton & Signals \cite{memmesheimer2020gimme} & \results{utdmhad_inter_sensor_inertial_skeleton_signals} \\ \midrule
           Skeleton & Inertial & Compact & \results{utdmhad_inter_sensor_skeleton_inertial_sldml}   \\
           Inertial & Skeleton & Compact & \results{utdmhad_inter_sensor_inertial_skeleton_sldml} \\
            \bottomrule
		\end{tabular}
	\end{center}

	\label{tab:oneshot_utdmhadintersensor}
\end{table}

\begin{table}[tb]
    \vspace{0.06in}
	\caption{one-shot action recognition results on the Simitate dataset.}
	\begin{center}
        \small
		\begin{tabular}{ccr}
% 			\toprule
             \#Train Classes  & \#Val Classes &  Accuracy [\%]\\
            \toprule
            22                 & 4       & \results{simitate_22_4} \\ \midrule
            18                 & 4         & \results{simitate_18_4} \\ 
            14                 & 4        & \results{simitate_14_4} \\  % 'val': {'epoch': 13, 'accuracy': 0.9322033898305084}})
            10                 & 4        & \results{simitate_10_4} \\  %'val': {'epoch': 1, 'accuracy': 0.9096045197740112}}
            \midrule
            % 22                 & 4       & 92.1 \\ %  'val': {'epoch': 13, 'accuracy': 0.9209039548022598}}) 
            18                 & 8         & \results{simitate_18_8} \\ %'val': {'epoch': 20, 'accuracy': 0.7918918918918919}})
            14                 & 12        & \results{simitate_14_12} \\ %'val': {'epoch': 3, 'accuracy': 0.5598739495798319}})
            10                 & 16        & \results{simitate_10_16} \\  %'val': {'epoch': 16, 'accuracy': 0.3624260355029586}})
            % Ours                                         & \textbf{88.1}
            %INFO:root:defaultdict(<class 'dict'>, {'samples': {'epoch': 10, 'accuracy': 0.4867513037981352}, 'val': {'epoch': 10, 'accuracy': 0.4867513037981352}})
            \bottomrule
		\end{tabular}
	\end{center}

	\label{tab:oneshot_simitate_results}
\end{table}

\subsection{Results}
\label{ssec:results}

On the NTU RGB+D 120 dataset we compare against the proposed baseline APSR by Liu \andothers \cite{liu2019ntu}. \tabname \ref{tab:oneshot_results} shows the results with an auxiliary set size of 100 action classes and a evaluation set size of previously unseen 20 action classes. Our proposed approach performs \ntuoneshotimpro\% better than the first follow up \cite{liu2019ntu} and \ntuoneshotimprosecond\% better than the second follow up \cite{liu2017skeleton}. \fignamelong \ref{fig:result_figs} \textit{(a)} and \tabname \ref{tab:ontshot2_results2} shows results for an increasing amount of auxiliary classes (100 auxiliary classes and 20 evaluation classes are considered as the standard protocol). Overall our approach performs better as the baseline on all conducted auxiliary set experiments. Interestingly to note is the high accuracy with 60 auxiliary classes and that the 20 additional classes added in the 80 classes auxiliary set added confusion. With 60 classes our approach performs 9.8\% better than the baseline approach with a same amount of auxiliary classes.
Further, our approach performs better, with just 60\% of the training data, than the first follow up with the full amount of auxiliary classes.
With only 40\% of the training data, our approach performs comparably good as the second and third follow up. These experiments strongly highlight the quality of the learned metric.
% Especially with a lower and higher amount of auxiliary classes our metric learning based approach performs better. 
% In comparison to the baseline the accuracy of our approach increased more linearly by increasing the amount of auxiliary classes. 
In \figname \ref{fig:umap_embedding} we show UMAP \cite{mcinnes2018umap} visualizations that give an insight about the discriminative capabilities. Distances in embedding space capture the amount of identities well. This is the case for the three top clusters containing the actions \textit{(grab other person's stuff, take a photo of other person} and \textit{hugging other person)}. The two clusters at x-axis around -7.5 correspond to the actions \textit{arm circles} and \textit{throw}, suggesting that actions with clear high joint-relevance can also be clustered well. The most bottom  cluster corresponds to the class \textit{falling} and supports this hypothesis. In the left we have a quite sparse cluster reflecting highly noisy skeleton sequences from multiple classes. Mainly sequences with multiple persons, especially with close activity like hugging, resulted in noisy data.
\tabname \ref{tab:oneshot_results_ntu_ablation} gives an ablation study showing the influence of the loss weighting and the underlying triplet mining approach. A multisimilarity miner \cite{wang2019multi} with a metric loss yields the best results fort one-shot action recognition on the NTU 120 dataset. In our ablation study the loss weighting with $\alpha 1.0, \beta 0.0$ yielded the best results for both miners.

The UTD-MHAD dataset was used to show the generalization capabilities of the proposed approach across different modalities. By considering a signal level action representation we could compare skeleton and inertial results and also perform multimodal, inter-joint  and inter-sensor experiments. \fignamelong \ref{fig:result_figs} \textit{(b)} shows the effect on the resulting one-shot accuracy with increasing auxiliary sets. Interesting to note is that in this experiment series we could observe that not necessarily a higher amount of classes used for training will lead to a higher accuracy. This was the case for our experiments on inertial data, where training on only three classes shows a more similar action embedding. This observation could not be transferred to the skeleton experiments on this dataset. 
The selection of auxiliary classes used for training should be well chosen. Adding more classes does not necessarily mean higher similarity in the embedding but can also add more confusion.
Our inter-joint experiments yielded more transferable embeddings by training on data from the wrist and validating on the leg as shown in \tabname \ref{tab:oneshot_utdmhadinterjoint}.  This holds true for our conducted experiments, but we do want not exclude the possibility of finding a subdivision of the wrist auxiliary set that results in a higher transferable embedding. 
A key-insight among our experiments is that balanced classes for training and testing yielded mostly higher accuracy for lower dimensional modalities like IMU (see \tabname \ref{tab:oneshot_utdmhad}) and motion capturing (see \tabname \ref{tab:oneshot_simitate_results}). This is especially visible in our inter-joint experiments as shown in \tabname \ref{tab:oneshot_utdmhadinterjoint}. In comparison, our experiments applied to skeleton sequences benefited from more auxiliary classes (see \tabname \ref{tab:oneshot_results}, \ref{tab:oneshot_utdmhad} and \figname \ref{fig:result_figs} \textit{(a,c)}). 
The conducted fusion experiments, by the the concatenation of skeleton sequences and inertial measurements show good performance in some experiments. The lower performing modality can also negatively impact the performance. This observation suggest to add sensor confidences into the approach as a future research direction. Sensor data fusion on an signal level by a single stream architecture remains an interesting and functional alternative to multi-stream architectures.
\begin{figure}[th] \centering$
  \vspace{0.06in}
  \begin{array}{cc}
      \includegraphics[width=.45\linewidth]{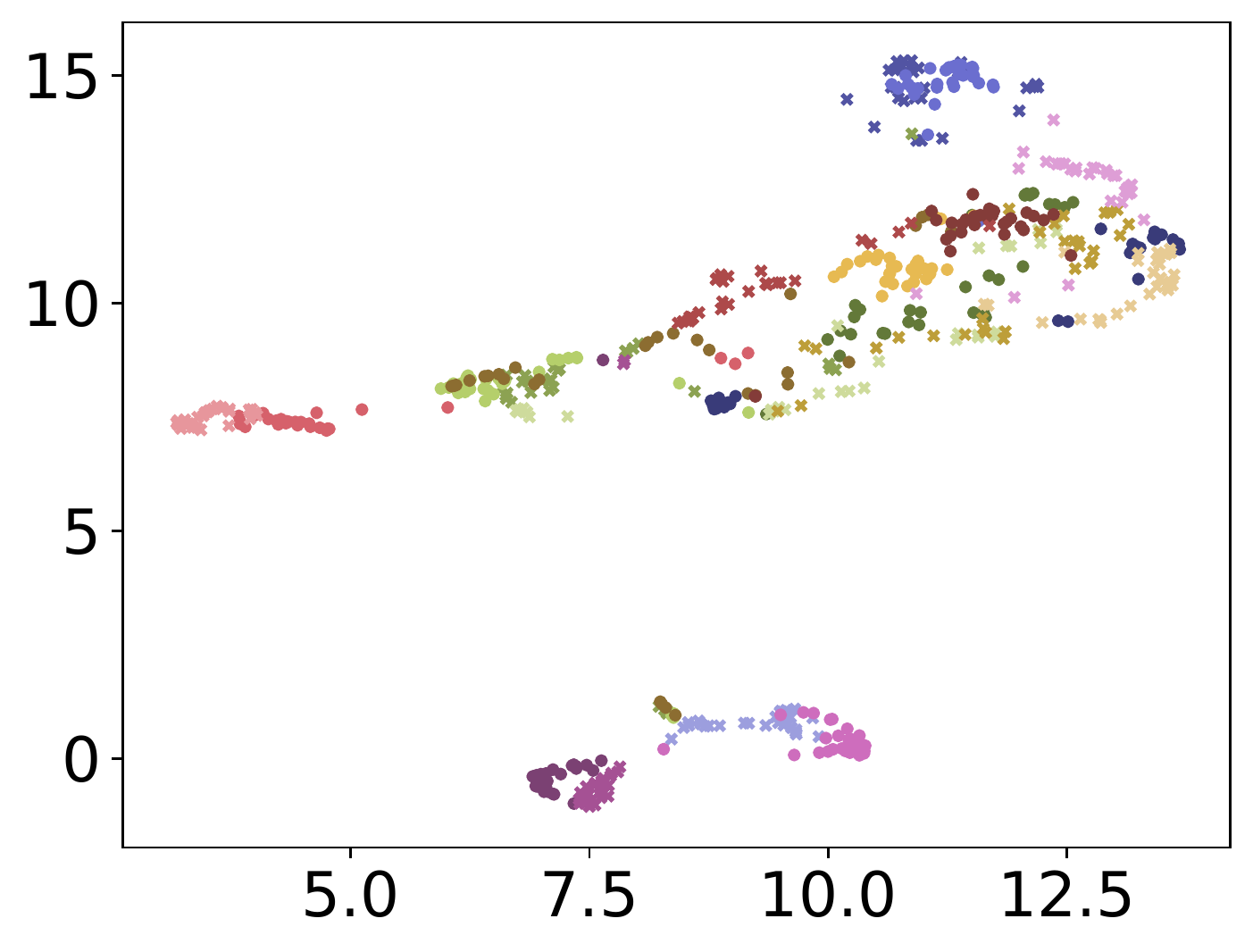} &
      \includegraphics[width=.45\linewidth]{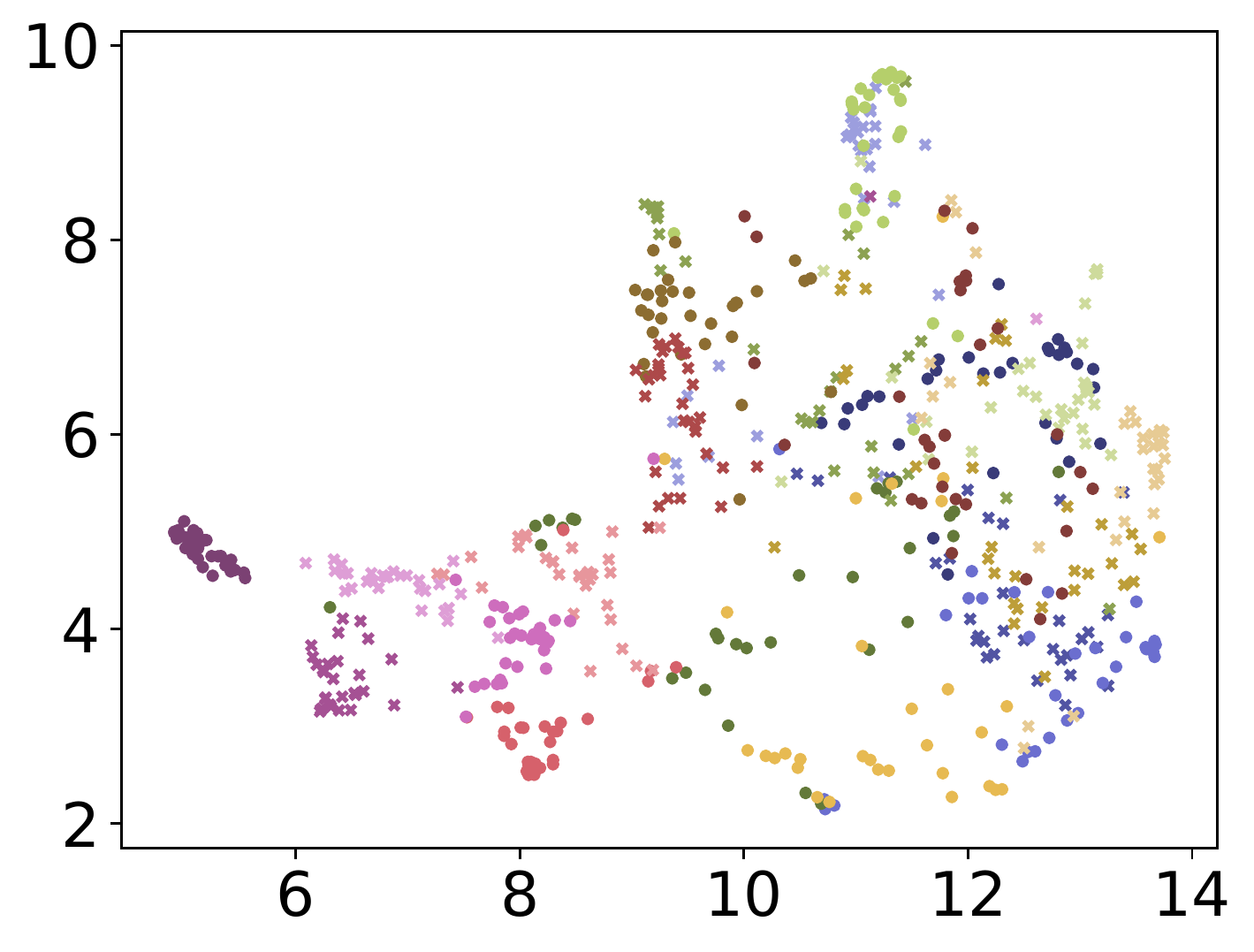} \\
      (a) & (b) 
  \end{array}$
  \caption{UMAP embedding visualization for inter-modal experiments. Trained on skeleton inertial measurements and validated on skeleton (\textit{a}). Trained on skeleton sequences and validated on inertial measurements (\textit{b}).}
  \label{fig:umap_embedding_intermodal}
\end{figure}
In our inter-sensor experiments we used all actions from one modality as auxiliary set and evaluated the other modality with a single reference sample. Results for this experiment are given in \tabname \ref{tab:oneshot_utdmhadintersensor} and \figname \ref{fig:umap_embedding_intermodal} visualizes the corresponding UMAP embeddings. The resulting one-shot recognition for inertial to skeleton performs by a large margin better (+17.4\%) better than the reverse direction using our novel representation. In approximately 40.5\% of the time an action trained on a different data modality on the UTD-MHAD dataset could be recognized with just one reference sample.
The signal representation \cite{memmesheimer2020gimme} generalize better in this aspect.
Overall the inter-modal experiments show the flexibility of our proposed approach but are subject for further improvement. We observed that the inertial measurements have a relation to the arm and hand movements of the skeleton, which explains the good transferability across the modalities.

% \textcolor{red}{Fusion experiment description}

Finally, we evaluated our approach on the Simitate dataset with motion capturing data. Results are given in \tabname \ref{tab:oneshot_simitate_results} and \figname \ref{fig:result_figs} {\textit(c)}. The amount of classes is comparable to the one from the UTD-MHAD dataset. The effects of the auxiliary set reduction are en par with the experiments conducted on the UTD-MHAD dataset. Therefore, the proposed approach transfers also good to motion capturing data. The class-distances from the motion capturing experiments are higher in embedding space then the ones gathered by the inertial experiments (see \figname \ref{fig:result_figs} {\textit(b)} \& {\textit(c)}).

\section{Conclusion}
We presented a one-shot action recognition approach by employing a signal level representation in conjunction with metric learning using a triplet margin loss. 
By considering a representation on a signal level, our approach remains flexible across different sensor modalities like skeleton, inertial measurements, motion capturing data or the fusion of multiple modalities. 
Our approach allows one-shot recognition on all of the modalities we experimented with and further indicated to serve as a flexible framework for inter-joint and even inter-sensor experiments. 
The novel inter-sensor experiments, by training on one modality and inferring on an unknown modality, can potentially shape future evaluation protocols.
We evaluate our approach on three different, publicly available, datasets. Most importantly, we showed an improvement of the current state-of-the-art for one-shot action recognition on the large scale NTU RGB+D 120 dataset. To show the transfer capabilities, we also verified our results using the UTD-MHAD dataset for skeleton and inertial data and the Simitate dataset for motion capturing data. Inter-joint experiments show inertial sensor attached to the wrist and the leg from the UTD-MHAD dataset.  On the UTD-MHAD dataset inter-sensor experiments between the skeleton and inertial data were executed. 
We found that more classes used during training for lower variate sensor data like IMUs and motion capturing systems do not necessarily improve the one-shot recognition accuracy in our experiments.  
A good selection of training classes and a balanced training and validation set
improved results across all modalities. 

\bibliographystyle{./IEEEtran}
\bibliography{references}

% \newpage
% \section{Author Response}
% \input{authors_response}

\end{document}